\pdfoutput=1

\documentclass[11pt]{article}

\usepackage{acl}

\usepackage{times}
\usepackage{latexsym}

\usepackage[T1]{fontenc}

\usepackage[utf8]{inputenc}

\usepackage{microtype}

\usepackage{inconsolata}
\usepackage{graphicx}
\usepackage{url}
\usepackage{enumitem}
\usepackage{easyReview}

\usepackage{multirow}
\usepackage{booktabs}
\usepackage{amsmath}

\title{\textit{Strength Lies in Differences!} Improving Strategy Planning for Non-collaborative Dialogues via Diversified User Simulation}

\author{
Tong Zhang$^{\spadesuit\heartsuit}$, \quad
Chen Huang$^{\spadesuit\heartsuit}$, \quad
Yang Deng$^{\diamondsuit}$, \quad
Hongru Liang$^{\spadesuit\heartsuit}$, \quad \\
\textbf{Jia Liu}$^{\clubsuit}$, \quad
\textbf{Zujie Wen}$^{\clubsuit}$, \quad
\textbf{Wenqiang Lei}$^{\spadesuit\heartsuit}$\thanks{ \quad Corresponding author.}, \quad
\textbf{Tat-Seng Chua}$^{\star}$
\\
${\spadesuit}$ Sichuan University \quad ${\diamondsuit}$ Singapore Management University \\ 
${\clubsuit}$ Ant Group, China \quad ${\star}$ National University of Singapore \\
${\heartsuit}$ Engineering Research Center of Machine Learning and Industry Intelligence, \\Ministry of Education, China \\
\{scu.zhangtong, huangc.scu\}@gmail.com \quad \{lianghongru, wenqianglei\}@scu.edu.cn \\
\{jianiu.lj,  zujie.wzj\}@antgroup.com \quad ydeng@smu.edu.sg \quad chuats@comp.nus.edu.sg
}

\begin{document}
\maketitle

\begin{abstract}
We investigate non-collaborative dialogue agents, which are expected to engage in strategic conversations with diverse users, for securing a mutual agreement that leans favorably towards the system's objectives. 
This poses two main challenges for existing dialogue agents: 1) The inability to integrate user-specific characteristics into the strategic planning, and 2) The difficulty of training strategic planners that can be generalized to diverse users.
To address these challenges, we propose \textsc{Trip} to enhance the capability in tailored strategic planning, incorporating a user-aware strategic planning module and a population-based training paradigm. Through experiments on benchmark non-collaborative dialogue tasks, we demonstrate the effectiveness of \textsc{Trip} in catering to diverse users.

\end{abstract}

\section{Introduction}
Non-collaborative dialogues, such as negotiation \cite{he2018decoupling} and persuasion \cite{wang-etal-2019-persuasion}, occur when the agent and user hold conflicting interests \cite{deng-etal-2023-goal, deng-rethinking, lei2022interacting}.
Typically, both parties need to employ various strategies to achieve an agreement favorable to themselves \cite{keizer-etal-2017-evaluating, zhang-etal-2023-towards-effective, zhan2024let}.
As user resistance varies depending on the agent's strategies \cite{shi2019build, dutt2021resper}, \textbf{it is imperative for the agent to perform strategic planning tailored to diverse users}. Relying on a one-size-fits-all strategy can leave the agent vulnerable to others taking advantage due to its lack of adaptability and flexibility \cite{yang2021improving, deng-towards, xu2023language}.

Recent efforts have resorted to large language models (LLMs) as dialogue agents to perform non-collaborative tasks \cite{deng2023prompting, fu2023improving, zhang2023ask}.
They aim to guide the response of LLMs through mixed-initiative prompts \cite{chen-etal-2023-controllable, deng2023prompting, zhang2023ask} or incorporating an external strategy planner \cite{yu-etal-2023-prompt, deng2023plug}. 
However, these initiatives has been criticized regarding its performance in real-world scenarios \cite{deng2023plug, kwon2024llms}, where users have various non-collaborative strategies. 
We attribute this outcome to the neglect of two crucial aspects: 
1) \textbf{Existing methods fail to incorporate explicit user-specific characteristics into their strategic planning}, instead relying solely on the conversational history.
Importantly, by creating informative representations of individual users, agents can adapt their behaviors and devise tailored strategies \cite{jang2020bayes, yang2021improving}.
2) \textbf{Their training paradigm fails to generate strategic planners that generalize well to diverse users}. 
Their paradigms are oversimplified, relying on a single user simulator for interactive training. This simulator is restricted in generating varied non-collaborative behaviors, often exhibiting a focus on prioritizing user contentment \cite{zhang2023explaining, durmus2023towards, bianchi2024well}. 
Essentially, agents trained in this manner are accustomed to engage with a single user exclusively, leading to rigidity and obstinacy when encountering new users with different interaction behaviors \cite{wang2023does, safdari2023personality}.

To provide more evidence for the above analysis, we establish an evaluation protocol, which situates diverse user simulators with varying non-collaborative behaviors.
We investigate the limitations of current LLM-based dialogue agents on strategic planning (cf. Section \ref{sec3} for details). 
The evaluation results clearly demonstrate that existing agents struggle to tailor their strategies for diverse users, leading to sub-optimal performances.
This limitation compromises the practical utility of these agents, both in functioning as a successful agent in conversational AI and in providing social skills training in pedagogy.
\textbf{The key challenges lie in making dialogue agents aware of diverse non-collaborative user behaviors and devising tailored strategies for individual users.}

To tackle these challenges, we design a simple yet effective method, called \textbf{\textsc{Trip}}, to improve LLMs' capability in \underline{T}ailored st\underline{R}ateg\underline{I}c \underline{P}lanning.
\textsc{Trip} includes a user-aware strategic planning module and a population-based training paradigm. 
Specifically, the strategic planning module incorporates user-specific characteristics into strategic planning using the Theory-of-Mind (ToM) \cite{premack1978does, wimmer1983beliefs}. 
This involves analyzing users' mental states and future possible actions during interactions to understand their interests 
\cite{yang2021improving, chawla2023social}.
Moreover, instead of relying on a solitary user simulator, our population-based training paradigm promotes the adaptation of the strategic planning module to various users, achieved by training it with more diverse user simulators. 
Each simulator is equipped with extensive sets of non-collaborative strategies and role-playing personas \cite{chen2024oscarsaitheatersurvey}. As such, \textsc{Trip} essentially manipulates the experience of the dialogue agent, enabling it to recognize the importance of tailoring strategies for individual users. 
Our key contributions are concluded below:
\begin{itemize}[leftmargin=*]
    \item We emphasize the significance of tailoring strategies for diverse users in non-collaborative dialogues. We verify the inadequacies of current LLM-based dialogue agents in this aspect.
    \item We propose \textsc{Trip} to achieve tailored strategic planning, which includes a user-aware strategic planning module and a population-based training paradigm.
    \item We conduct experiments on benchmark non-collaborative dialogue tasks (i.e., negotiation and persuasion). Our findings suggest that \textsc{Trip} is proficient in catering to diverse users using tailored strategies, consistently outperforming baselines across different tasks.
\end{itemize}

\section{Related Work}
\label{related_work}
Our research is closely tied to the strategic planning and training paradigms to address the non-collaborative tasks in the era of LLMs. We provide a literature review and highlight our differences.

\noindent \textbf{Strategic planning for non-collaborative dialogues}. 
Recent researches have introduced various methods based on LLMs to enhance their effectiveness in strategic planning. 
These methods can be categorized into two types: 
1) \textit{Developing stimulus prompts to unleash the potential of LLMs}.
\cite{chen-etal-2023-controllable} validate the effectiveness of using mixed-initiative prompts to tackle proactive dialogue challenges. 
\cite{deng2023prompting} and \cite{zhang2023ask} encourage LLMs to engage in self-reflection to plan their next actions. 
\cite{fu2023improving} employ self-play simulations to iteratively refine strategic planning by soliciting feedback from other LLMs.
Nonetheless, as highlighted by \cite{deng2023plug}, the effectiveness of these approaches is impeded by non-trainable parameters.
2) \textit{Equipping LLMs with an external strategy planner}. 
The planner is capable of generating prompts at each turn, providing nuanced, instance-specific guidance and control over LLMs.
This could be integrated using methods like Monte Carlo Tree Search \cite{yu-etal-2023-prompt} or a plug-in model \cite{deng2023plug}, which can be fine-tuned for improving the strategic planning capability without affecting the functionalities of LLM-powered dialogue agents. 
However, these methods still struggle to achieve promising results due to their inability to integrate user-specific characteristics into their strategic planning.
Complementary to \cite{deng2023plug}, our work investigates the importance of tailored strategic planning by modeling user-related characteristics explicitly.

\noindent \textbf{Training paradigms for non-collaborative dialogues}.
Current training paradigms involve the dialogue agent interacting with a single user simulator to enhance its strategic planning capabilities.
In specific, \cite{chawla-etal-2023-selfish} build a user simulator that mimics human-human dialogue data in a supervised manner, while \cite{yu-etal-2023-prompt, deng2023plug} resort to a role-playing LLM-based user simulator.
However, a single user simulator can only represent the behaviors of one or a type of users, potentially leading to the under-representation of other users' behaviors, as evidenced by \cite{liu2023one, shi2019build}.
Therefore, existing training paradigms fail to produce strategic planners that cater to diverse users with varying behaviors. 
In this paper, our work investigates the importance of tailored strategic planning by diversifying the user's behaviors using population-based training.

\section{Strategic Planning Evaluation}
\label{sec3}
We introduce a novel evaluation protocol to analyze the limitations of existing LLM-based dialogue agents and highlight their inability to handle users exhibiting various non-collaborative behaviors. 
The overall evaluation process is illustrated in Figure \ref{evaluation_environment}.
See more details of our evaluation protocol in Appendix \ref{app:evaluation_environment}.

\begin{figure}[t] 
\includegraphics[width=.48\textwidth]{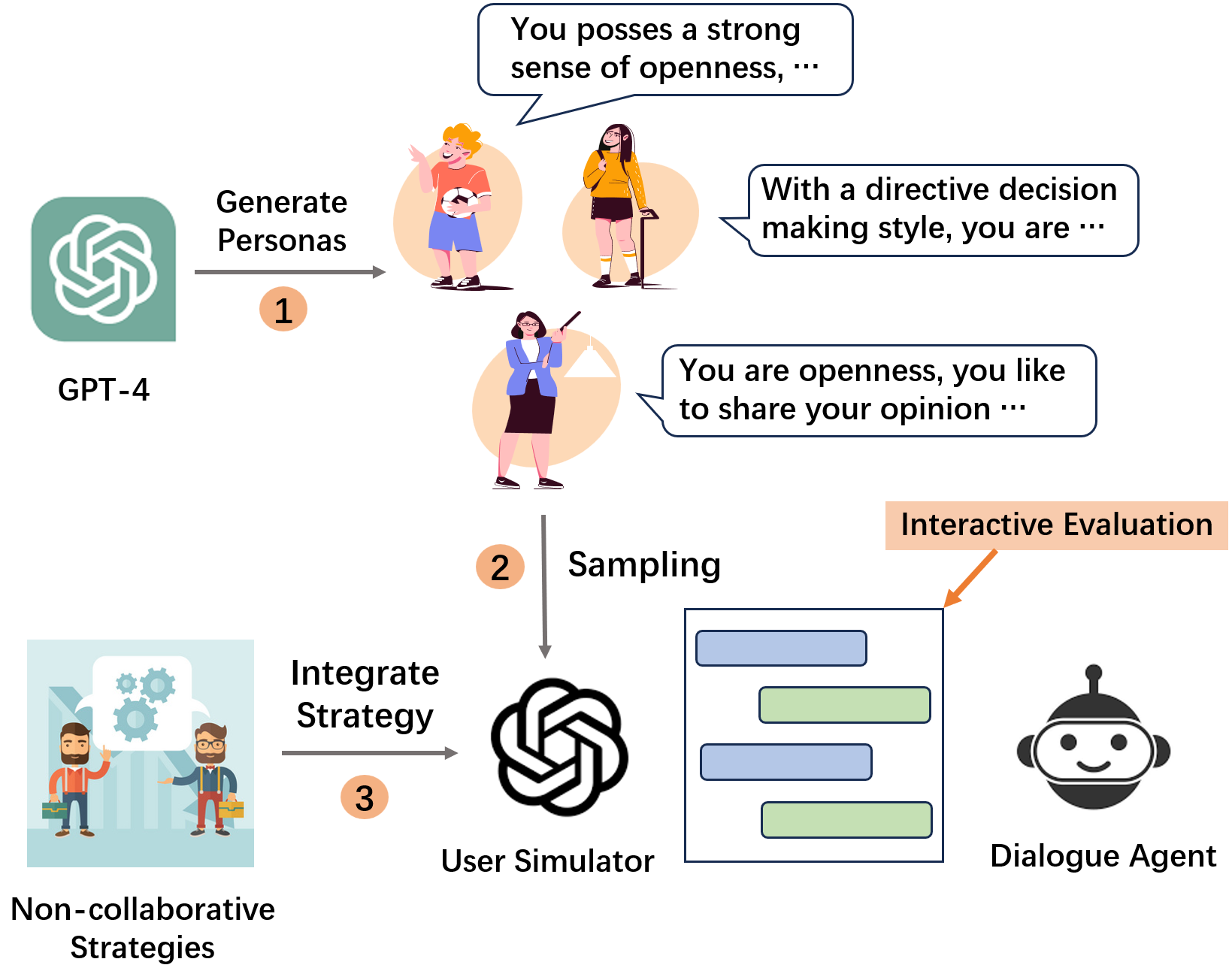} 
\centering
\caption{The overall evaluation process.} 
\label{evaluation_environment} 
\end{figure}

\subsection{Evaluation Setup}
\label{user_simulators}
\noindent \textbf{Evaluation Overview}.
The environment encompasses various synthetic user simulators showcasing diverse non-collaborative behaviors. 
In the evaluation process, each dialogue agent must interact with these simulators \cite{deng2023plug}.
During their interactions, the dialogue agent and user simulator alternate in employing strategies in their responses with the ultimate aim of maximizing their own self-interest. 
The interactions continues until the conversational goal is achieved or the maximum number of turns is reached. 
We gather these interactions and assess the agents performances.

\noindent \textbf{Baselines}.
We consider two representative baselines: \textit{Standard} agent (i.e., vanilla LLM without any modification) and \textit{PPDPP} agent \cite{deng2023plug}, which is current SOTA agent with a trainable external strategy planner\footnote{Notably, we also consider other existing dialogue agents in our main experiments.}.

\noindent \textbf{Diverse User Simulators}.
Our simulators are synthesized with non-collaborative behaviors, guided by their task-relevant personas. 
As evidenced by previous study \cite{deng2023survey, bianchi2024well, huang2024conceptevaluationprotocol}, LLMs are limited to demonstrate non-collaborative behaviors. To this end, we prompt non-collaborative behaviors explicitly into LLMs using the \textit{resisting strategies} that are designed to foil persuasion attempts \cite{fransen2015strategies, tian-etal-2020-understanding, dutt2021resper}.
Initially, we equip LLMs with different personas \cite{jiang2023personallm, zhou2023sotopia, zhang-etal-2023-towards-effective}, which are used to select non-collaborative behaviors from the set of \textit{resisting strategies}. 
Following \cite{wang-etal-2019-persuasion, jiang2024evaluating}, we consider two types of personas, including \textit{Big-Five Personality}\footnote{Openness, Conscientiousness, Extraversion, Agreeableness, and Neuroticism} \cite{goldberg1992development} and \textit{Decision-Making Styles}\footnote{Directive, Conceptual, Analytical, and Behavioral} \cite{scott1995decision}, together with LLM-generated cohesive description for each fine-grained persona.
Additionally, we employ \textit{resisting strategies} outlined by \cite{dutt2021resper} to direct the behavior of simulators.
Finally, our mixed-initiative role-play prompt for each agent includes the assigned persona, a set of resisting strategies, and conversation context. These elements aid in guiding user simulators to exhibit diverse non-collaborative behaviors.
In total, we develop 300 diverse user simulators for each evaluation task, representing 20 persona categories (i.e., Big-Five Personality $\times$ Decision-Making Styles).




\noindent \textbf{Evaluation Tasks}. In line with \cite{deng2023prompting, wang-etal-2019-persuasion}, we conduct experiments on two benchmark non-collaborative tasks: the price negotiation task, utilizing the test\footnote{Our data split follows the previous study \cite{deng2023plug, wang-etal-2019-persuasion}.} dataset of CraigslistBargain (CB) \cite{he2018decoupling} and the charity persuasion task, employing the test dataset of PersuasionForGood (P4G) \cite{wang-etal-2019-persuasion}. Notably, the dialogue agents play the role of buyer and persuader, respectively, to accomplish their goals.

\begin{table*}[ht]
\small
\tabcolsep=0.4cm
\centering
\resizebox{\linewidth}{!}{
\begin{tabular}{cl|ccc|cc}
\hline
\multicolumn{2}{c|}{\multirow{2}{*}{\textbf{Personas}}}            & \multicolumn{3}{c|}{\textbf{Price Negotiation}}         & \multicolumn{2}{c}{\textbf{Persuasion for Good}} \\
\multicolumn{2}{c|}{}                                              & SR$\uparrow$          & AT$\downarrow$           & SL\%$\uparrow$          & SR$\uparrow$             & AT$\downarrow$              \\ \hline
\multicolumn{1}{c|}{\multirow{5}{*}{Big Five}} & Openness          & 0.76$_{\textcolor{red}{\uparrow0.23}}$ & 6.66$_{\textcolor{red}{\uparrow0.63}}$ & 0.34$_{\textcolor{red}{\uparrow0.12}}$ & 0.47$_{\textcolor{red}{\uparrow0.34}}$    & 8.92$_{\textcolor{red}{\uparrow1.00}}$    \\
\multicolumn{1}{c|}{}                          & Conscientiousness & 0.69$_{\textcolor{red}{\uparrow0.25}}$ & 7.20$_{\textcolor{red}{\uparrow1.04}}$ & 0.27$_{\textcolor{red}{\uparrow0.06}}$ & 0.39$_{\textcolor{red}{\uparrow0.33}}$    & 8.90$_{\textcolor{red}{\uparrow1.10}}$    \\
\multicolumn{1}{c|}{}                          & Extraversion      & 0.74$_{\textcolor{red}{\uparrow0.16}}$ & 6.17$_{\textcolor{red}{\uparrow1.47}}$ & 0.39$_{\textcolor{red}{\uparrow0.15}}$ & 0.45$_{\textcolor{red}{\uparrow0.35}}$    & 8.73$_{\textcolor{red}{\uparrow1.25}}$    \\
\multicolumn{1}{c|}{}                          & Agreeableness     & 0.40$_{\textcolor{red}{\uparrow0.01}}\star$ & 6.82$_{\textcolor{red}{\uparrow0.71}}$ & 0.28$_{\textcolor{red}{\uparrow0.06}}$ & 0.18$_{\textcolor{red}{\uparrow0.12}}$    & 9.85$_{\textcolor{red}{\uparrow0.13}}\star$    \\
\multicolumn{1}{c|}{}                          & Neuroticism       & 0.31$_{\textcolor{blue}{\downarrow0.02}}\star$ & 6.81$_{\textcolor{red}{\uparrow1.12}}$ & 0.20$_{\textcolor{blue}{\downarrow0.02}}\star$ & 0.12$_{\textcolor{red}{\uparrow0.02}}\star$    & 9.78$_{\textcolor{red}{\uparrow0.14}}\star$    \\ \hline
\multicolumn{1}{c|}{\multirow{4}{*}{Decision}} & Analytical        & 0.37$_{\textcolor{red}{\uparrow0.04}}\star$ & 7.07$_{\textcolor{red}{\uparrow0.61}}$ & 0.26$_{\textcolor{red}{\uparrow0.06}}\star$ & 0.16$_{\textcolor{red}{\uparrow0.09}}$    & 9.43$_{\textcolor{red}{\uparrow0.56}}\star$   \\
\multicolumn{1}{c|}{}                          & Directive         & 0.41$_{\textcolor{red}{\uparrow0.05}}\star$ & 6.71$_{\textcolor{red}{\uparrow1.48}}$ & 0.18$_{\textcolor{blue}{\downarrow0.03}}\star$ & 0.12$_{\textcolor{blue}{\downarrow0.02}}\star$    & 9.31$_{\textcolor{red}{\uparrow0.62}}$    \\
\multicolumn{1}{c|}{}                          & Behavioral        & 0.78$_{\textcolor{red}{\uparrow0.25}}$ & 6.45$_{\textcolor{red}{\uparrow1.20}}$ & 0.39$_{\textcolor{red}{\uparrow0.16}}$ & 0.53$_{\textcolor{red}{\uparrow0.37}}$    & 8.94$_{\textcolor{red}{\uparrow1.04}}$    \\
\multicolumn{1}{c|}{}                          & Conceptual        & 0.77$_{\textcolor{red}{\uparrow0.23}}$ & 6.62$_{\textcolor{red}{\uparrow0.78}}$ & 0.42$_{\textcolor{red}{\uparrow0.17}}$ & 0.49$_{\textcolor{red}{\uparrow0.36}}$    & 9.02$_{\textcolor{red}{\uparrow0.94}}$    \\ \hline
\multicolumn{2}{c|}{Overall Performance}                           & 0.58$_{\textcolor{red}{\uparrow0.14}}$ & 6.72$_{\textcolor{red}{\uparrow1.01}}$ & 0.31$_{\textcolor{red}{\uparrow0.09}}$ & 0.32$_{\textcolor{red}{\uparrow0.23}}$    & 9.20$_{\textcolor{red}{\uparrow0.76}}$    \\ \hline

\end{tabular}
}
\caption{The performance of the \textit{PPDPP} dialogue agent testing across various personas of user simulators. \textcolor{red}{Red} (\textcolor{blue}{Blue}) indicates the increased (decreased) performance compared to \textit{Standard} dialogue agent. 
The symbol $\star$ indicates that this performance exhibits minimal variation, specifically within a 5\% range of the maximum value.
The effectiveness of \textit{PPDPP} varies significantly across different user personas.}
\label{preliminary_persona_results}
\end{table*}

\noindent \textbf{Evaluation Metrics}. 
\label{preliminary_metrics}
Following \cite{deng2023plug}, we consider three commonly used metrics: \underline{Success Rate} (SR), \underline{Average Turn} (AT) and \underline{Sale-to-List Ratio} (SL\%).
The SR measures effectiveness by the percentage of goal achievement within a maximum number of turns, while AT measures efficiency by the average number of turns required to achieve the goal.
As for the CB task, we additionally adopt the SL\% \cite{zhou2019augmenting} to determine the effectiveness of goal completion.
Formally, the SL\% is expressed as $(P_{deal} - P_{target}^{seller}) / (P_{target}^{buyer} - P_{target}^{seller})$, where $P_{deal}$ is the final deal price, $P_{target}^{buyer}$ and $P_{target}^{seller}$ are the target prices of both parties.
A higher SL\% represents the buyer gets more benefits from the deal.
If failing to reach a deal at the end, we set SL\% as 0.

\subsection{Experimental Findings}
\label{experimental_investigation}
We analyze the performances of existing dialogue agents across user simulators with various non-collaborative behaviors.
Specifically, we assess the advancements of \textit{PPDPP} compared to the \textit{Standard} agent.
As illustrated in Table \ref{preliminary_persona_results}, while \textit{PPDPP} shows a notable improvement in overall performance, it does not adapt well to users employing different non-collaborative strategies. 
Its effectiveness varies significantly among users with different personas, with its advantage over the \textit{Standard} not being significant in 17.77\% of cases (e.g., it increases SR by 0.02 for \textit{Analytical} in price negotiation.), and even performing worse than the \textit{Standard} in 8.88\% of cases (e.g., it decreases SR by 0.02 for \textit{Neuroticism} in price negotiation). This motivates the need for a dialogue agent to perform strategic planning tailored to diverse users\footnote{We find that other baselines also have similar issues, as detailed in Section \ref{eeeee}.}.

\begin{figure*}[ht] 
\includegraphics[width=\textwidth]{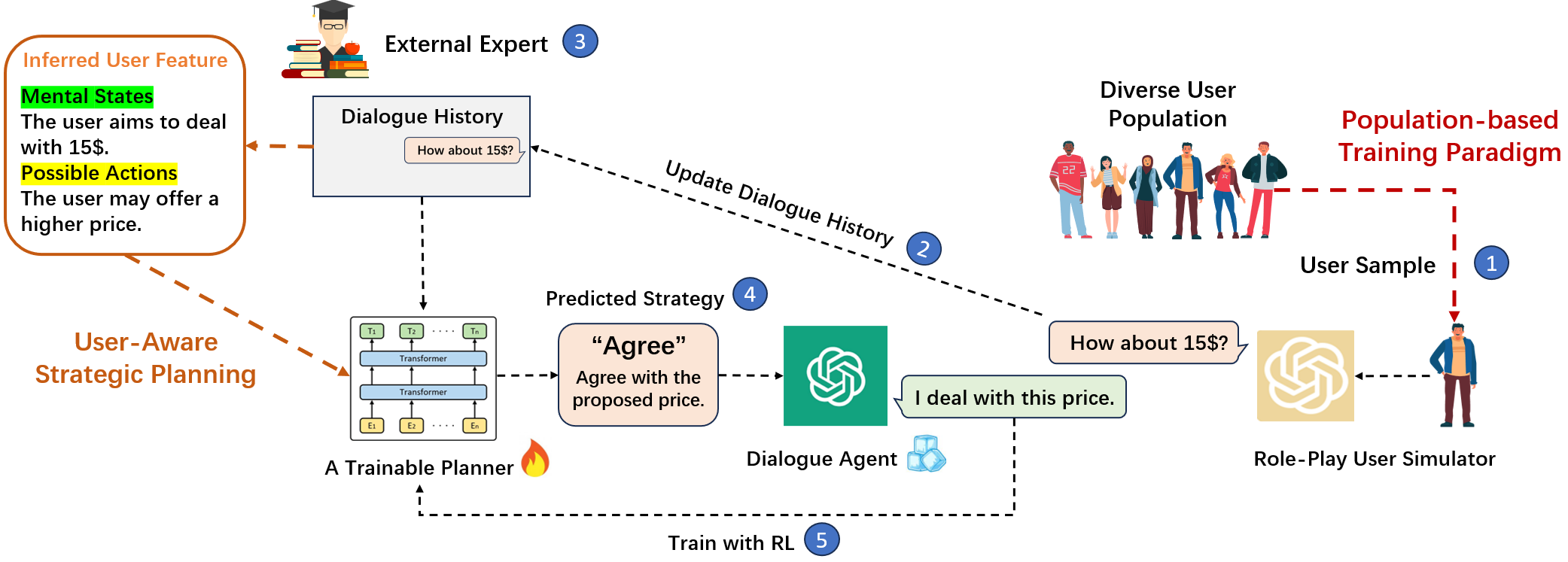} 
\centering
\caption{TRIP Overview. This method includes a user-aware strategic planning module (UASP) and a population-based training paradigm (PBTP). 
The UASP incorporates user-specific characteristics into strategic planning using the Theory-of-Mind (ToM). 
The PBTP diversifies training user simulators to promote agents' adaptation. We use numbers to indicate the overall process of TRIP.} 
\label{TRIP_overview} 
\end{figure*}

\section{\textsc{Trip}: Tailored Strategic Planning}
To enhance LLMs' tailored strategic planning, we propose an effective method \textsc{Trip}, which develops an external planner by modeling user characteristics and training with diverse user simulators.
As illustrated in Figure \ref{TRIP_overview}, our \textsc{Trip} includes a user-aware strategic planning module and a population-based training paradigm. 
The former aims to explicitly model user characteristics (e.g., mental states and future actions), while the latter incorporates diverse user simulators for training simultaneously. 

\subsection{User-Aware Strategic Planning}
\textsc{Trip} aims to explicitly infer user characteristics and then incorporate them into the strategic planning module, parameterized by a trainable BERT. 
In particular, building upon the advanced \textit{Theory-of-Mind} capability of LLMs \cite{sap2022neural, moghaddam2023boosting}, \textsc{Trip} captures users’ mental states and future possible actions during interactions to understand their interests and predicts how TRIP’s responses may influence them. In this case, mental states pertains to what they aim to accomplish, such as the target price or whether they will donate, while future actions relates to what the user is likely to discuss next \cite{hu2023enhancing, zhou2023far}.
Formally, given the dialogue history $D =(u_1^{sys},u_1^{usr},...,u_{t}^{sys},u_{t}^{usr})$, where $u_i^{sys}$ and $u_i^{usr}$ denote the $i$-th utterances of both parties and $t$ is the number of utterances, we feed the dialogue history $D$ into the LLM and prompt it to infer mental states $\mathcal{M}$ and future actions $\mathcal{F}$ in an open-ended manner, i.e., $P_{LLM}(\mathcal{M},\mathcal{F}|D)$.
Subsequently, we feed the \{$\mathcal{M},\mathcal{F}, D$\} into the strategy planner $\pi_\theta$ to predict the next strategy.
The output space of $\pi_\theta$ is a set of strategies\footnote{e.g., the elicitation of specific emotions to influence other.} pre-defined by \cite{deng2023plug, wang-etal-2019-persuasion}, each of them is attached with a pre-defined natural language instructions.

\subsection{Population-based Training Paradigm} 
Given that a single user simulator tends to favor limited behaviors while under-represents others \cite{shi2019build, liu2023one}, we explore training a dialogue agent using a set of user simulators employing different non-collaborative strategies to accommodate diverse users.
To achieve this, we propose a population-based reinforcement learning (RL) training paradigm, which aims to enhance the adaptability of a dialogue agent to new user groups by training with larger and more diverse populations \cite{charakorn2020investigating}. We offer a comprehensive explanation of this approach below.

\noindent \textbf{Population Setup}. 
Similar to Section \ref{user_simulators}, we build 40 diverse user simulators, each embodying a specific persona description.
We ensure an balanced representation of each persona category within our user simulators for population-based RL training.
We donate these simulators as $K = {k_1, k_2, ...k_{40}}$ 
During each iteration, we sample among $K$ using a distribution $p$, allowing the dialogue agent $S$ to interact with it.
The distribution $p$ is initialized based on the frequency of various personas.

\noindent \textbf{Reward Design}.
Following \cite{deng2023plug}, we prompt LLMs to judge the conversation progress at each turn and transform it into scalar rewards.
Specifically, in the negotiation task, we employ a separate GPT3.5 \cite{ChatGPT} to assess whether both parties have reached a deal.
In the persuasion task, we ask the GPT3.5-based user simulator to express its willingness to donation.
Our rewards are determined based on three situations: 
1) Successful goal achievement by the dialogue agent results in a significant positive reward, defined as 1.0 in the charity persuasion task and the value of SL\% in the price negotiation task.
2) Failure to achieve goals leads to a substantial negative reward of -1.0 for the dialogue agent. 
3) Furthermore, we assign a small negative reward (-0.1) per turn to penalize the lengthy conversation, which promotes the efficient goal achievement.


\noindent \textbf{Optimization}. 
During RL training, we maximize the expected reward of the strategy planner $\pi_\theta$ by utilizing the REINFORCE algorithm \cite{williams1992simple}: $\theta \leftarrow \theta - \alpha \nabla \log \pi_\theta R_t$,
where $\theta$ denotes the trainable parameter of the strategy planner, $\alpha$ denotes the learning rate, and $R_t$ is the total reward accumulating from turn $t$ to the final turn $T$: $R_t=\sum^T_{t'=t}\gamma^{T-t'}r_{t'}$, where $\gamma$ is a discount factor.


\begin{figure*}[ht] 
\includegraphics[width=\textwidth]{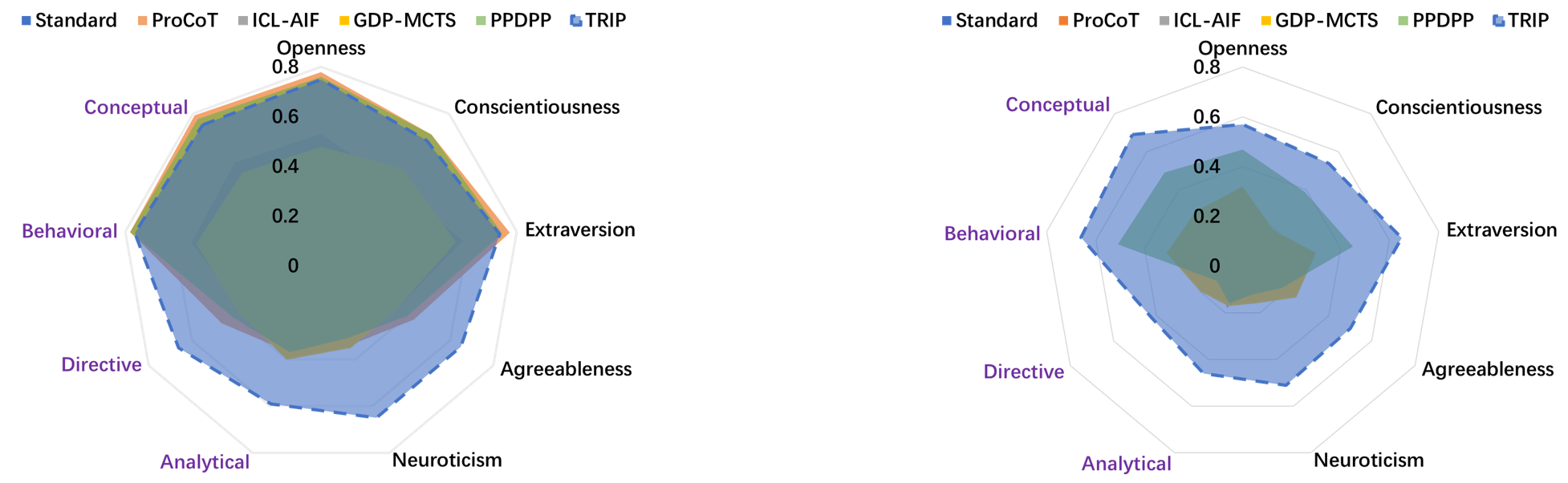}
\centering
\caption{The agents performance across various personas. We report their success rate on two tasks, namely price negotiation (\textit{Left}) and charity persuasion (\textit{Right}).
\textsc{Trip} achieves balanced improvements on all personas, significantly outperforming other agents by a considerable margin.
Due to limited space, we report other results using different metrics in Appendix \ref{app:radar_persona_results}.}
\label{Radar_SR} 
\end{figure*}

\section{Experiments}
\label{eeeee}
This sections aims to evaluate the effectiveness of our \textsc{Trip}, following the evaluation protocol proposed in Section \ref{user_simulators}. 
We initially report the overall performances of dialogue agents in Section \ref{overaaa}.
Next, we conduct an in-depth analysis to reveal the tailored strategies of \textsc{Trip} in Section \ref{indepth_analysis}.
Finally, we perform ablation studies in Section \ref{ablation_study} to 
sort out the performance variation of different user awareness and training population, and find a dominant predictor for the tailored strategic planning.

\noindent \textbf{LLM-based baselines}.
We consider LLM-based dialogue agents with two types of strategic planning modules, as discussed in Section \ref{related_work}. 
1) \underline{Prompt-based planning}, including \textit{Standard}, \textit{ProCoT} \cite{deng2023prompting} and \textit{ICL-AIF} \cite{fu2023improving}, which use mixed-initiative prompts, CoT, and AI feedback to select next strategies, respectively.
2) \underline{External strategy planners}, including \textit{GDP-MCTS} \cite{yu-etal-2023-prompt} and \textit{PPDPP} \cite{deng2023plug}, which utilize Monte Carlo Tree Search and a trainable plug-in for determining next-step strategies, respectively. 
Note that all baselines fail to model user-specific characteristics explicitly and are trained using one user simulator. 
Implementation details are presented in Appendix \ref{app:implementation}.

\noindent \textbf{Evaluation Metrics}. 
We use the same automatic metrics mentioned in section \ref{preliminary_metrics}.
Furthermore, we conduct human evaluation to assess the practical effectiveness of these dialogue agents. 
See more details of human evaluation in Appendix \ref{app:human_evaluation}.

\subsection{Overall Performance}
\label{overaaa}
\begin{figure}[ht] 
\includegraphics[width=.48\textwidth, height=0.32\textwidth]{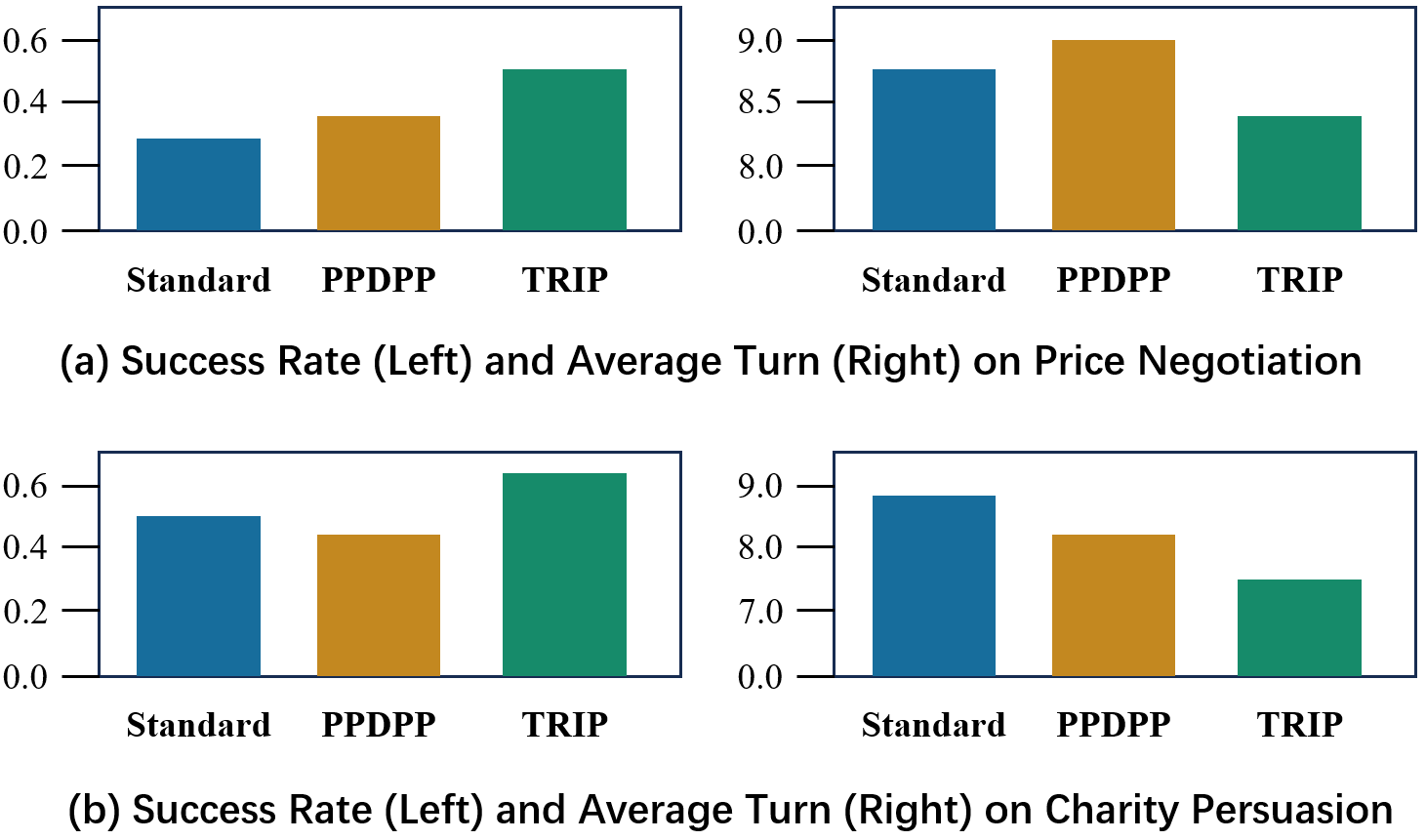} 
\centering
\caption{Human Evaluation Results. \textsc{Trip} shows a high practical utility to deal with real users.} 
\label{Human_Evaluation} 
\end{figure}

We evaluate the overall and fine-grained performance of all agents using automatic metrics in Table \ref{overall_results} and Figure \ref{Radar_SR}. Additionally, we report human evaluation in Figure \ref{Human_Evaluation} to gauge their performance during interactions with real users.

\noindent \textbf{\textsc{Trip} is a promising method for achieving effective non-collaborative strategies tailored for diverse users}. 
As illustrated in Table \ref{overall_results}, \textsc{Trip} significantly outperforms all the baselines with a noticeable margin across two tasks.
It not only efficiently achieves the conversational goal (less AT) but also effectively accomplishes tasks (higher SR and higher SL\%).
Moreover, as depicted in Figure \ref{Radar_SR}, \textsc{Trip} shows balanced improvements across different user personas, significantly outperforming other agents by a substantial margin, in contrast to the biased improvements of \textit{PPDPP} in Section \ref{experimental_investigation}.
This suggests that \textsc{Trip} is capable of generating strategies that generalize well to diverse users.
This also implies that the behavior pattern pf a single LLM-based user simulator is limited in scope.
Moreover, our human evaluation results in Figure \ref{Human_Evaluation} show our \textsc{Trip} largely outperform the \textit{Standard} and \textit{PPDPP} when interacting with real users.
Notably, we observed that \textit{PPDPP} does not consistently surpass the \textit{Standard} approach across the two tasks. For instance, while it achieves a higher success rate in the negotiation task, it necessitates more interaction rounds.
This evidences the effectiveness and practical utility of our proposed \textsc{Trip}.

\begin{table}[ht]
\renewcommand\arraystretch{1.1}
\tabcolsep=0.4cm
\centering
\resizebox{\linewidth}{!}{
\begin{tabular}{l|ccc|cc}
\toprule
\multicolumn{1}{c|}{\textbf{\large{Agents}}} & \multicolumn{3}{c|}{\textbf{Price Negotiation}} & \multicolumn{2}{c}{\textbf{Persuasion for Good}} \\
                                     & SR$\uparrow$            & AT$\downarrow$         & SL\%$\uparrow$         & SR$\uparrow$                  & AT$\downarrow$                \\ \hline
Standard                            & 0.4444        & 7.73       & 0.2222       & 0.0930              & 9.96              \\
ProCoT                              & 0.6040        & 7.62       & 0.2307       & 0.1833              & 9.90              \\
ICL-AIF                             & 0.3411        & 8.42       & 0.2503       & 0.1667              & 9.91              \\
GDP-MCTS                                 & 0.4444        & 7.63       & 0.2401       & 0.2466              & 9.74              \\
PPDPP                              & 0.5855        & 6.72       & 0.3144       & 0.3233              & 9.20              \\ \midrule
TRIP (\textit{Ours})                            & \textbf{0.6888}        & \textbf{6.34}       & \textbf{0.4096}       & \textbf{0.5533}              & \textbf{8.51}              \\ 
\bottomrule
\end{tabular}
}
\caption{Overall evaluation. \textsc{Trip} is promising for achieving effective non-collaborative strategies.}
\label{overall_results}
\end{table}

\begin{figure*}[ht] 
\includegraphics[width=\textwidth]{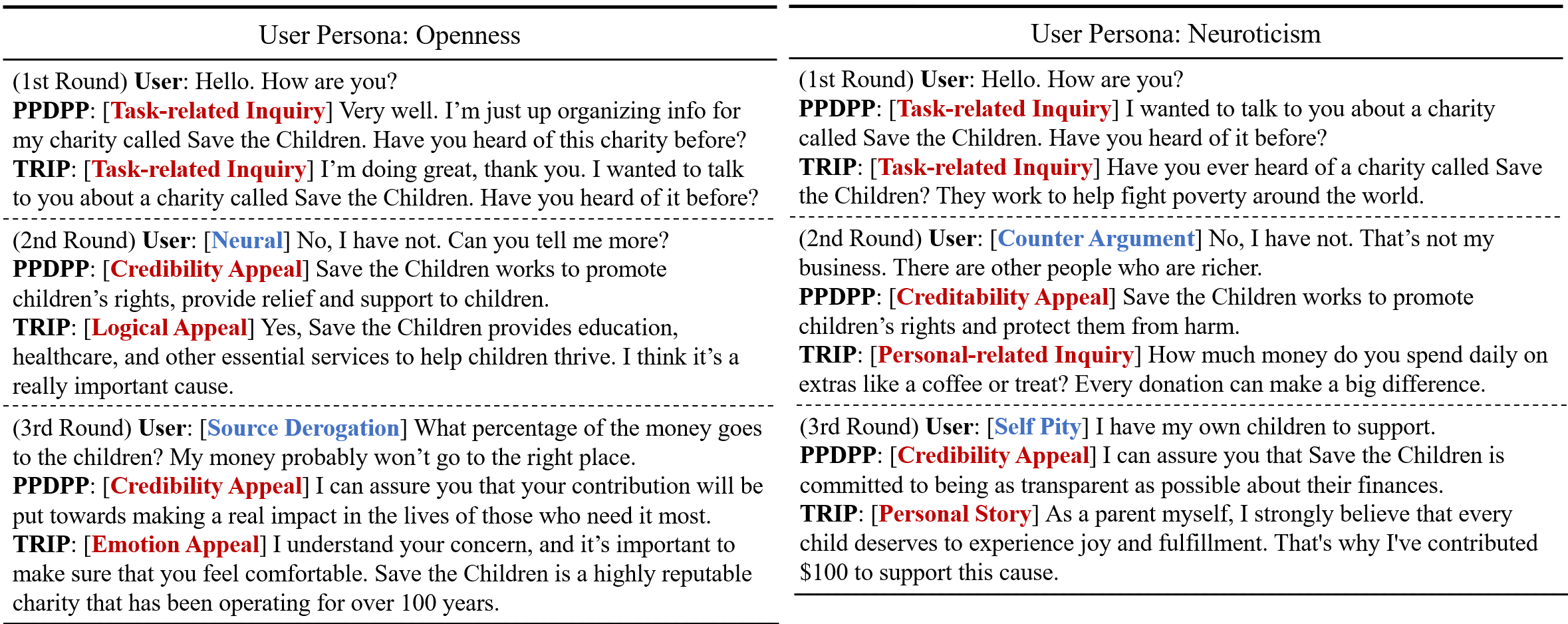} 
\centering
\caption{Case study on the charity persuasion task (Top-3 conversation rounds). The user resisting strategies and agent strategies are marked in \textcolor[RGB]{68,114,196} {bleu} and \textcolor[RGB]{192,0,0} {red} respectively. 
While \textit{PPDPP} repeats its strategy usage pattern to different user types, \textsc{Trip} effectively tailor its strategies for different users. When dealing with the\textit{Openness} persona (\textit{Left}), \textsc{Trip} introduces the charitable organization and evoke specific emotions to sway users' decision. Conversely, in addressing the  \textit{Neuroticism} persona (\textit{Right}), \textsc{Trip} tends to discuss personal experiences related to charity and employs reasoning persuade the user.
} 
\label{case_study} 
\end{figure*}

\subsection{Strategy Analysis}
\label{indepth_analysis}
In this section, we analyze the effectiveness of our \textsc{Trip} in tailored strategic planning.
Specifically, in each user interaction, we gather the strategies employed by each agent at every turn and combine them in a sequential order to form a strategy sequence. Then, we compare the strategy sequences employed by different agents.
We utilize BERT \cite{devlin2018bert} and the t-SNE method \cite{van2008visualizing} to encode each strategy sequence into an embedding vector.
Subsequently, we use the Euclidean distance measure to calculate the average distance between any two strategy sequences used by agents with the same persona, as well as the average distance between any two strategy sequences used by agents with different personas. 
This is akin to the metrics (i.e., the Intra-Class and Inter-Class analysis) used in the metric learning community \cite{Roth_2019_ICCV} and we term them as the Intra-Persona and Inter-Persona. The results are shown in Table \ref{cluster_persona}.


\begin{table}[t]
\centering
\small
\begin{tabular}{l|c|c}
\toprule
Models   & \textbf{Intra-Persona}$\downarrow$ & \textbf{Inter-Persona}$\uparrow$ \\ \midrule
Standard & 24.93         & 13.51         \\
ProCoT    & 21.37         & 15.65         \\
ICL-AIF    & 22.84         & 15.33         \\
GDP-MCTS    & 20.72         & 16.09         \\
PPDPP    & 19.37         & 17.28         \\ \midrule
TRIP (\textit{Ours})     & \textbf{16.14}         & \textbf{20.26}         \\ \bottomrule
\end{tabular}
\caption{The strategy distribution of different agents. The Intra-Persona metric donates the average distance for a particular persona. The Inter-Persona metric donate the average distance for different personas. \textsc{Trip} achieves the best performance, showcasing its effectiveness in devising tailored strategies for diverse users.}
\label{cluster_persona}
\end{table}

\noindent \textbf{\textsc{Trip} demonstrates a greater awareness of population dynamics, resulting in reduced variance across specific user simulators.}
As shown in Table \ref{cluster_persona}, \textsc{Trip} achieves the lowest Intra-Persona and the highest Inter-Persona.
This indicates that the strategy sequences of \textsc{Trip} exhibit similarity when interacting with users sharing the same personas and non-collaborative behaviors.
Also, these sequences are distinct when compared to users with different personas.
This further reveals that \textsc{Trip} holds advantages in devising tailored strategies for diverse users.

For better understanding, we present a case study in Figure \ref{case_study} and examine the strategy sequence employed by \textit{PPDPP} and \textsc{Trip} in an charity persuasion task.
Specifically, \textit{PPDPP} repeats its strategy usage pattern to different user types, briefly using of credentials and citing organizational impacts to establish credibility and earn the persuadee’s trust.
In contrast, \textsc{Trip} demonstrates a deeper understanding of the users and provides more tailored strategies.
When dealing with the \textit{Neuroticism} persona, \textsc{Trip} tends to discuss personal experiences related to charity and employs reasoning persuade the user.
Conversely, in addressing the \textit{Openness} persona, \textsc{Trip} introduces the charitable organization and evoke specific emotions to sway users' decision.
The strategy sequence used by \textsc{Trip} is believed to be more persuasive, as demonstrated by \cite{barford2016openness, wang-etal-2019-persuasion}, stating that the \textit{Openness} users are inclined to embrace novelty and be easily influenced by emotions, while the \textit{Neuroticism} users are more likely to be influenced by others' personal experiences.
In this regard, we believe that these strategic differences may provide valuable insights for the future research on the non-collaborative dialogues.



\subsection{Ablation Study}
\label{ablation_study}
This section aims to sort out the performance variation of different user awareness and training population.
To analyze the effectiveness of each design, we consider the following variants of \textsc{Trip}.
\begin{itemize}[leftmargin=*, , itemindent=0.05cm, itemsep=-4pt]
    \item \textbf{\textsc{Trip}\textsubscript{w/o POP}}: We eliminate the population-based training approach from \textsc{Trip} and instead have \textsc{Trip} engage with a single fixed LLM-based user simulator for training, without any specific role-playing persona.

    \item \textbf{\textsc{Trip}\textsubscript{w/o UA}}: We remove the user-aware strategic planning module, and only takes the conversation history as inputs to plan next strategies. 
    \item \textbf{\textsc{Trip}\textsubscript{w/ 10 POP}}: It utilizes 10 personas for population training, each simulator is randomly selected from a pool of 20 persona categories.

    \item \textbf{\textsc{Trip}\textsubscript{w/ 10 POP \& w/o UA}}: In this variant, we remove the user-aware strategic planning module from \textit{\textsc{Trip} w/ 10 POP}.

\end{itemize}

We summarize the overall performance of each model variation Table \ref{ablation_results}. 
Based on these results, we draw the following observations:

\begin{table}[t]
\renewcommand\arraystretch{1.15}
\tabcolsep=0.4cm
\centering
\resizebox{\linewidth}{!}{
\begin{tabular}{l|ccc|cc}
\toprule
\multicolumn{1}{c|}{\textbf{Models}} & \multicolumn{3}{c|}{\textbf{Price Negotiation}} & \multicolumn{2}{c}{\textbf{Persuasion for Good}} \\
                                     & SR$\uparrow$            & AT$\downarrow$         & SL\%$\uparrow$         & SR$\uparrow$                  & AT$\downarrow$                \\ \hline
\textbf{\textsc{Trip}}                & \underline{0.6888}        & \underline{6.34}       & \textbf{0.4096}       & \textbf{0.5533}              & \textbf{8.51}              \\ \hline
\textbf{\textsc{Trip}\textsubscript{w/o UA}}                    & \textbf{0.6988}        & 6.38       & 0.3881       & 0.5133              & 8.69              \\

\textbf{\textsc{Trip}\textsubscript{w/o POP}}                            & 0.5766        & 7.00       & 0.3505       & 0.4400              & 8.95              \\
\textbf{\textsc{Trip}\textsubscript{w/ 10 POP \& w/o UA}}                    & 0.6377        & 6.73       & 0.3543       & 0.4700              & 8.79              \\
\textbf{\textsc{Trip}\textsubscript{w/ 10 POP}}                & 0.6700        & \textbf{6.12}       & 0.3537       & 0.4733              & 8.72              \\ \hline
\textbf{PPDPP}                                & 0.5855        & 6.72       & 0.3144       & 0.3233              & 9.20              \\ \bottomrule

\end{tabular}
}
\caption{The evaluation results of ablation study. The user-aware strategic planning module and population-based training are effective to improve agents and complement each other.}
\label{ablation_results}
\vspace{-3mm}
\end{table}

\noindent \textbf{User-aware strategic planning and population-based training paradigm are both effective to produce tailored strategic planning}.
Specifically, compared to \textsc{Trip}\textsubscript{w/o UA}, we note \textsc{Trip} improves the persuasion success rate (0.3233 $\rightarrow$ 0.4400) and the deal benefit SL\% (0.3144 $\rightarrow$ 0.3505).
This suggest that incorporating user mental states and future actions can assist the agent in developing more effective strategies.
Notably, this variant slightly decreases the deal success rate (0.6988 $\rightarrow$ 0.6888).
This can be attributed to the fact that deeply modeling user characteristics may inadvertently decrease the seller's willingness to engage in the deal, as the focus is on maximizing one's own benefits.
Moreover, compared to \textsc{Trip}\textsubscript{w/o POP}, we observe that \textsc{Trip} yield positive improvements across all metrics, such as  significant increase in SL\% (0.3505 $\rightarrow$ 0.4096).
This demonstrates that diversifying the behaviors of training user simulators effectively improves the agent's performance.

\begin{figure}[t] 
\includegraphics[width=.48\textwidth]{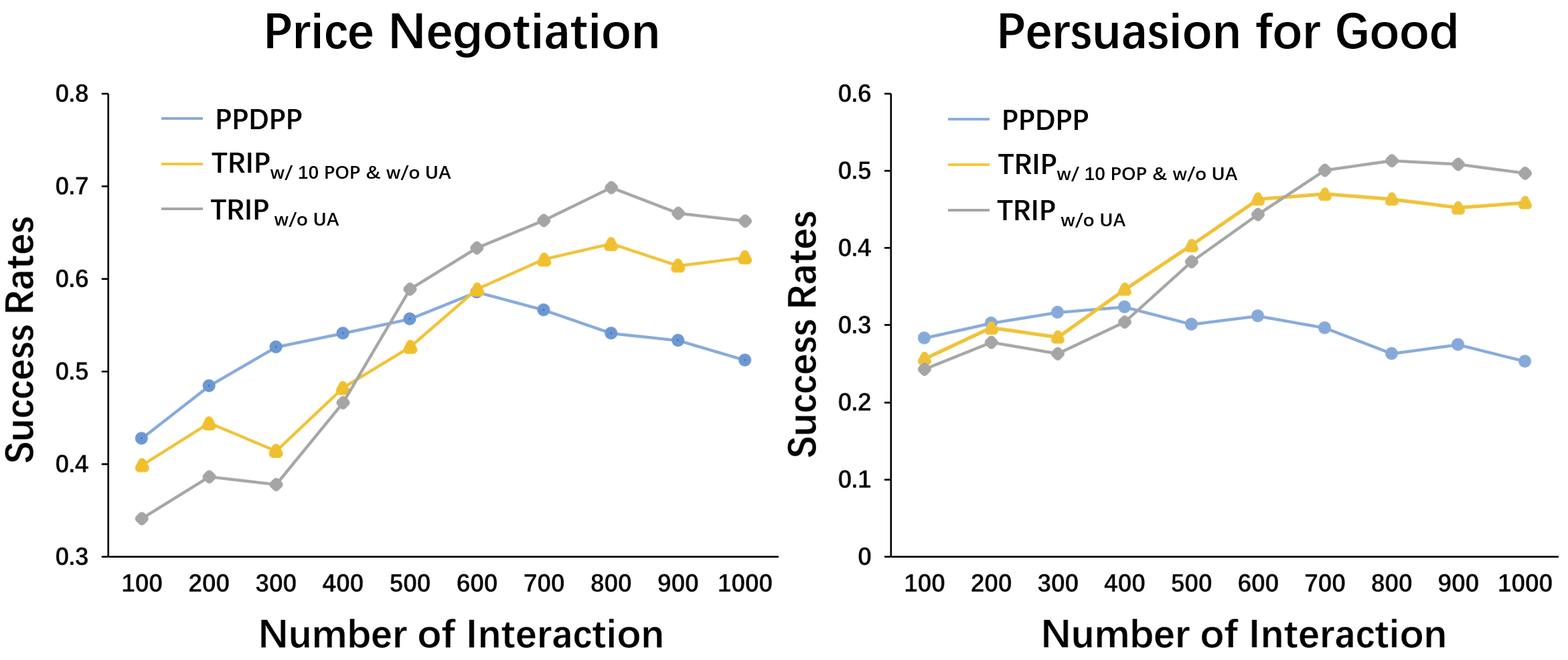} 
\centering
\caption{The test performance of different number of training user simulators. \textit{PPDPP} converges easily but has a limited upper bound in terms of performance.} 
\label{enhancing_user_diversity} 
\vspace{-2mm}
\end{figure}

\noindent \textbf{Diverse training populations is more beneficial to improve the adaptability of dialogue agents, but it may also present additional training challenges}.
As shown in Table \ref{ablation_results}, compared to \textsc{Trip}\textsubscript{w/o UA} and \textsc{Trip}\textsubscript{w/o POP}, we find that diverse training populations is more important for \textsc{Trip}'s superiority. 
Moreover, we find that \textsc{Trip}\textsubscript{w/o UA} demonstrates higher performances than \textsc{Trip}\textsubscript{w/ 10 POP \& w/o UA} and \textit{PPDPP} (i.e., A single fixed user simulator).
To provide a detailed understanding of the impact of the number of training user simulators, we present their test performance of in 1000 training interactions, as depicted in Figure \ref{enhancing_user_diversity}.
Particularly, during the initial 400 interactions, we observe that \textsc{Trip}\textsubscript{w/o UA} and \textsc{Trip}\textsubscript{w/ 10 POP \& w/o UA} exhibit slower convergence compared to \textit{PPDPP}.
This suggests that not keeping the training user simulator fixed can introduce instability in the initial training phase, as also noted in \cite{lewis2017deal}.
However, beyond 500 interactions, the training process of \textsc{Trip}\textsubscript{w/o UA} stabilizes, leading to a significant performance enhancement, surpassing the other two agents.
Additionally, it is observed that \textit{PPDPP}'s performance declines after specific interactions (e.g., 600 in price negotiation), suggesting that extensive interactions with a single user simulator cannot consistently enhance agents' performance.

\section{Conclusion}
In this study, we investigate the inadequacies of current LLM-based dialogue agents in catering in diverse non-cooperative users. To address this, we propose \textsc{Trip}, a method designed to tailor strategic planning for non-collaborative dialogues. The idea behind our \textsc{Trip} is simple, involving a user-aware strategic planning module and a population-based training paradigm. Experimental results across diverse users demonstrate the superior effectiveness and efficiency of \textsc{Trip}. We consider our work as laying the groundwork for enhancing the adaptability and flexibility of non-cooperative dialogue agents in the era of LLMs. Moving forward, we plan to further explore the potential of population-aware agents in reducing the capital expenditure associated with training and coaching novice agents.

\section*{Limitations}

In this section, we discuss the limitations of this work from the following perspectives:

\noindent \textbf{Sensitivity of Prompts.} 
Similar to other studies on prompting LLMs \cite{deng2023prompting}, 
the evaluation results are expected to be influenced by the prompts.
Following \cite{deng2023plug}, we employ the mixed-initiative format to formulate our prompts, as it offers stability and control.
The impact of prompts and their optimality present important areas of investigation within LLMs, calling for exploration in future studies.

\noindent \textbf{Limited Non-collaborative Tasks.}
We only conduct our experiments on the two non-collaborative dialogue tasks (i.e., price negotiation and charity persuasion) due to their status as classic and widely-recognized benchmarks \cite{deng2023prompting, chawla2023social}.
In the future, we plan to apply our proposed \textsc{Trip} in a broader range of non-collaborative dialogue scenarios \cite{zhang-etal-2024-clamber, zhou2023sotopia}.

\section*{Acknowledgements}
This work was supported in part by the National Natural Science Foundation of China (No. 62272330 and No. 62206191);
in part by the Natural Science Foundation of Sichuan (No. 2023NSFSC0473); in part by the Fundamental Research Funds for the Central Universities (No. 2023SCU12089 and  No. YJ202219); in part by the Singapore Ministry of Education (MOE) Academic Research Fund (AcRF) Tier 1 grant (No. MSS24C004).

\bibliography{anthology,custom}

\appendix

\section{Details about Evaluation Protocol}
\label{app:evaluation_environment}

\subsection{Building User Simulators}
Due to the significant human labor required for real-user evaluations \cite{huang-etal-2023-reduce}, our experiments utilize user simulators instead.

\subsubsection{Persona Generation}
We prompt GPT4 \cite{OpenAI2023GPT4TR} to generate diverse user personas by selecting attributes from two persona types, namely \textit{Big-Five Personality} and \textit{Decision-Making Styles}.
Specifically, We allow GPT-4 to choose an attribute for each persona type, resulting in attribute-based user personas comprised of two fields, each containing a distinct attribute value.
The prompt we use is provided in Table \ref{prompt:persona generation}.
In total, we create 20 attribute-based user personas and ensure that the number of each attribute is balanced.
We then prompt GPT4 to rephrase these attribute-based personas into 300 cohesive persona descriptions.
The prompt we use is provided in Table \ref{prompt:persona rephrase}.

\subsubsection{Non-collaborative Behavior Prompting}
We leverage the \textit{resisting strategies} outlined in \cite{dutt2021resper} as users' non-collaborative behaviors.
We provide the detailed explanations of these resisting strategies in Table \ref{tab:framework}.
We design detailed instructions and incorporate these \textit{resisting strategies} with their explanations into our user simulator prompting.

\subsubsection{Comprehensive Prompting}
By incorporating the persona description and resisting strategies, we construct comprehensive prompts for our user simulators.
Specifically, our prompt includes two parts: task background and conversation history. 
In the task background, we guide LLMs to role-play their assigned personas with a set of role-play instructions and resisting strategies.
We provide the comprehensive user simulator prompts across two tasks in Table \ref{prompt:comprehensive user simulator CB} and \ref{prompt:comprehensive user simulator P4G}.

\subsection{Evaluation Tasks}
Following \cite{bianchi2024well, deng2023plug}, we consider two classic tasks as our evaluation scenarios, including price negotiation \cite{he2018decoupling} and charity persuasion \cite{wang-etal-2019-persuasion}.
The price negotiation task involves open-ended price negotiations where a buyer influences the seller towards a reasonable price, while the seller aims to maximize their own profit. 
The charity persuasion task involves asymmetric interactions guided by a persuader who endeavors to persuade the other party to make a charitable donation.
Our evaluation is based on these two tasks, requiring the evaluated dialogue agents to take on the roles of buyer and persuader, respectively, in order to achieve their goals. 
To support our evaluations, we adopt the test dataset of CraigslistBargain \cite{he2018decoupling} and PersuasionForGood \cite{wang-etal-2019-persuasion}, making use of their pre-annotated background information to streamline our assessment process.
For the negotiation task, the background information includes item details and the desired price of each party. 
For the persuasion task, it involves determining if the individual being persuaded initially intends to make a donation.
These background information serve as specific scenarios for our evaluation.




\begin{table}[!h]
\small
\centering
\resizebox{\linewidth}{!}{
\begin{tabular}{l|c|c}
\toprule
\textbf{CB}    & \textbf{Seller (User)} & \textbf{Buyer (Agent)} \\ \midrule
Target prices & 285\textdollar & 142\textdollar \\ \midrule
Item & \multicolumn{2}{c}{A skillfully lugged and elegantly pantographed road bike} \\ \midrule
Goals & Maximize the price & Minimize the price \\ \midrule
Ending condition & \multicolumn{2}{c}{When either party accepts} \\ \midrule
Max. \# of turns & \multicolumn{2}{c}{10 rounds of interaction} \\
\bottomrule
\end{tabular}
}
\caption{The evaluation scenario of price negotiation. This case is selected from the validate set of CraigslistBargain Dataset \cite{he2018decoupling}.}
\label{tab:sellerandbuyer:game:structure}
\end{table}

\begin{table}[!h]
\centering
\small
\resizebox{\linewidth}{!}{
\begin{tabular}{l|c|c}
\toprule
\textbf{P4G}    & \textbf{Persuader (Agent)} & \textbf{Persuadee (User)} \\ \midrule
Charity info & \multicolumn{2}{c}{It works to help fight poverty around the world}   \\ \midrule
Goals & Convince the persuadee to donate & Foil the persuasion \\ \midrule
Ending condition & \multicolumn{2}{c}{When the persuadee agree to donate.} \\ \midrule
Max. \# of turns & \multicolumn{2}{c}{10 rounds of interaction} \\
\bottomrule
\end{tabular}
}
\caption{The evaluation scenario of charity persuasion.}
\label{tab:ultimatum:game:structure}
\end{table}

\begin{table*}[t]
\small
\centering

\begin{tabular}{p{0.16\textwidth} p{0.37\textwidth}p{0.38\textwidth}}
\toprule
Resisting Strategy & Persuasion (P4G) & Negotiation (CB)\\ \midrule

Source Derogation & Attacks/doubts the organisation's credibility. & Attacks the other party or questions the item.\\ \hline

Counter Argument & Argues that the responsibility of donation is not on them or refutes a previous statement.  & Provides a non-personal argument/factual response to refute a previous claim or to justify a new claim. \\ \hline

Personal Choice & Attempts to saves face by asserting  their personal preference such as their choice of charity and their choice of donation. & Provides a personal reason for disagreeing with the current situation or chooses to agree with the situation provided some specific condition is met.\\ \hline

Information Inquiry & Ask for factual information about the organisation for clarification or as an attempt to stall. & Requests for clarification or asks additional information about the item or situation. \\ \hline

Self Pity & Provides a self-centred reason for not being able/willing to donate at the moment. & Provides a  reason (meant to elicit sympathy) for disagreeing with the current terms.\\ \hline

Hesitance & Attempts to stall the conversation by either stating they would donate later or is currently unsure about donating.  & Stalls for time and is hesitant to commit; specifically, they seek to further the conversation and provide a chance for the other party to make a better offer. \\ \hline

Self-assertion & Explicitly refuses to donate without even providing a factual/personal reason. & Asserts a new claim or refutes a previous claim with an air of finality/ confidence.\\ \hline

Others & Do not explicitly foil the persuasion attempts.  & Do not explicitly foil the negotiation attempts. \\ \bottomrule
\end{tabular}
\caption{The resisting strategies for P4G and CB tasks.}
\label{tab:framework}
\end{table*}

\begin{table}[t]
\centering
\small
\begin{tabular}{lcccc}
\toprule  
&\multicolumn{2}{c}{Single-turn}&\multicolumn{2}{c}{Multi-turn}\\
\cmidrule(lr){2-3}\cmidrule(lr){4-5}
Setting &Natural   & Useful &Natural   & Useful\\
\midrule
Human & 18\% & 20\% & 15\% & 22\%\\  
\textsc{Trip} & \textbf{45\%} & \textbf{42\%} & 34\% & 31\%\\  
Tie & 37\% & 38\% & \textbf{51\%} &\textbf{48\%}\\  
\bottomrule
\end{tabular}
\caption{Comparison on user simulators and real users. The Cohen’s Kappa between annotators is 0.67. }
\label{tab:user_sim}
\end{table}

\subsection{Reliability Analysis}
Prior to conducting the interactive evaluation, we validate the reliability of using LLMs as user simulators that demonstrate non-collaborative behaviors.
Following the approach described in \cite{deng2023plug}, we engage 5 human experts in conversations with two groups, including our diverse user simulators and 10 real users across two evaluation tasks.
We collect 50 dialogues from each group and evaluate the user responses in both single-turn and multi-turn open-ended conversations. 
The evaluation focuses on the naturalness and utility of the generated responses in these conversation settings. Naturalness refers to the fluency and human-like nature of the responses, while utility indicates their consistency with the role instructions and non-collaborative behaviors. 
We employ two annotators to conduct pairwise evaluations by rating "Win/Tie/Lose" between the two samples.
As shown in Table \ref{tab:user_sim}, the user simulators exhibit a notably superior performance compared to real users, particularly when it comes to the naturalness of responses in multi-turn conversations, which showcases the impressive language generation capabilities inherent in LLMs. 
Furthermore, even compared with human-annotated dialogues, the GPT3.5-based simulator shows competitive performance. 
These results validate the reliability of adopting GPT3.5 as the user simulator.

\subsection{Interactive Evaluation Protocol}
During the evaluation, each dialogue agent must engage with these simulators \cite{deng2023plug}.
During interactions, the dialogue agent and user simulator alternate in employing strategies in their responses with the ultimate aim of maximizing their own self-interest. 
The interactions continues until the conversational goal is achieved or the maximum number of turns T (i.e., T is set to 10 for both tasks) is reached.
To determine goal achievement, we utilize AI feedback to assess whether the task goal has been reached.
Specifically, in price negotiation task, we employ a separate GPT3.5 (i.e., $LLM_{rwd}$) to assess whether both parties have reached a deal. 
We prompt $LLM_{rwd}$ to generate feedback for the binary question “Have they reached a deal?”. 
If the output of $LLM_{rwd}$ indicates that both parties have reached an agreement, we consider this as goal achievement. 
In charity persuasion task, we additionally prompt the user simulator to express his willingness to make a donation at the end of each turn.
In particular, we query the user simulator "Would you be interested in donating to Save the Children?".
If the feedback is positive, we regard this as goal achievement.
Conversely, if the goal is not achieved, the interaction continues.

Due to the subjectivity of the planning outcome as well as the variance of the LLM-generated output,
we follow a common practice \cite{wang2022self, deng2023plug} to alleviate these issues by sampling the decoded sequences l (i.e., l is set to 10 for both tasks) times.

\section{Implementation Details}
\label{app:implementation}

\subsection{TRIP Implementation Details}
\label{app:trip_implementation}

\subsubsection{Theory-of-Mind}
We leverage the strong Theory-of-Mind capability of GPT3.5 to infer the mental states and user future actions during interaction.
The prompt we use is provided in Table \ref{prompt:tom_CB} and \ref{prompt:tom_P4G}.

\subsubsection{Strategy Prompting}
Here, we present the dialogue agent strategies utilized in our experiments.
Initially, we outline the strategies along with their explanations for two tasks in Table \ref{tab:negotiate_strat} and \ref{tab:persuasion_strat}.
We then offer a comprehensive overview of our \textsc{Trip} prompting in Table \ref{prompt:trip_agent_CB} and \ref{prompt:trip_agent_P4G}.

\subsubsection{Supervised Fine-Tuning}
\label{app:sft_planner}

We initialize our strategy planner by imitating human-human dialogue datasets in CraigslistBargain and PersuasionForGood through supervised fine-tuning (SFT).
In specific, we adopt the strategy annotations in the train dataset to support our SFT.
we optimize the strategy planner by minimizing the cross-entropy loss between the predicted strategy $y_i$ and the human annotated strategy $\hat{y_i}$:
\begin{equation}
\small
    \mathcal{L}_{CE} = -\frac{1}{m}\sum_{i=1}^{m}\left[ y\log\hat{y_i} + (1-y_i)\log(1-\hat{y_i}) \right] \nonumber
\end{equation}
Regarding the training hyper-parameters, we set the batch size 16 and the learning rate 6e-6, and utilize the AdamW optimizer with a weight decay of 0.01.
We save the checkpoint based on the best performance at the validation set.

\subsubsection{Online RL Training}
After SFT, we optimize our strategy planner through REINFORCE algorithm.
In specific, our training involves 1000 episodes, with a learning rate of 1e-6, a discount factor 0.999, and the maximum conversation turn of each episode 10. 
All the training experiments are run on a server equipped with 4 Tesla V100 GPUs.

\subsection{Baselines Implementation Details}
\label{app:baseline_implementation}
We implement the existing LLM-based dialogue agents by following previous works. 

\noindent \textbf{Standard}: simply prompts LLMs to chat with users using task instructions without considering any dialogue strategy.

\noindent \textbf{ProCoT}: we follow \cite{deng2023prompting} and prompt LLM to analyze the dialogue status and plan next strategy, and then generate a response based on the planned strategy.
We provide its prompt design in Table \ref{prompt:procot_agent}.

\noindent \textbf{ICL-AIF}: we follow \cite{fu2023improving} and prompt another GPT3.5 for verbal feedback, offering suggestions to the dialogue agent upon completion of an interaction.
Our implementation involves presenting three suggestions at the conclusion of each interaction, while ensuring that only the most recent 20 suggestions are retained to prevent indefinite expansion.
The prompt we use is provided in Table \ref{prompt:icl_aif_agent}.

\noindent \textbf{GDP-MCTS}: we follow \cite{yu-etal-2023-prompt} and implement open-MCTS to help LLM for strategic planning. 
This method is originally proposed for charity persuasion dialogues. 
In order to further accommodate the price negotiation applications, we just need to modify the task instruction and the role-playing description.

\noindent \textbf{PPDPP}: we follow \cite{deng2023plug} and adopt the BERT\footnote{https://huggingface.co/google-bert/bert-base-uncased} model \cite{devlin2018bert} as our external planner.
We implement \textit{PPDPP} based on the training details provided in the original paper.
We have made adjustments to the task instructions and role-playing descriptions, adapting them for use in the context of charity persuasion.

\section{Human Evaluation}
\label{app:human_evaluation}
Inspired by \cite{yu-etal-2023-prompt}, we conduct interactive human evaluation using the LegoEval platform \cite{li2021legoeval} with crowdworkers on Amazon Mechanical Turk.
We primarily sought to evaluate \textsc{Trip} against two competitive baselines (i.e., \textit{Standard} and \textit{PPDPP}).
In specific, we hire 20 crowd-workers with varying personas to converse with our three agents based on the price negotiation and charity persuasion tasks.
After conversations, we collect 50 dialogues for each agent and calculate their performances using the same metrics mentioned in Section \ref{preliminary_metrics}.

\section{More Experimental Results}
\label{app:radar_persona_results}
\begin{figure}[ht] 
\includegraphics[width=.48\textwidth, height=0.32\textwidth]{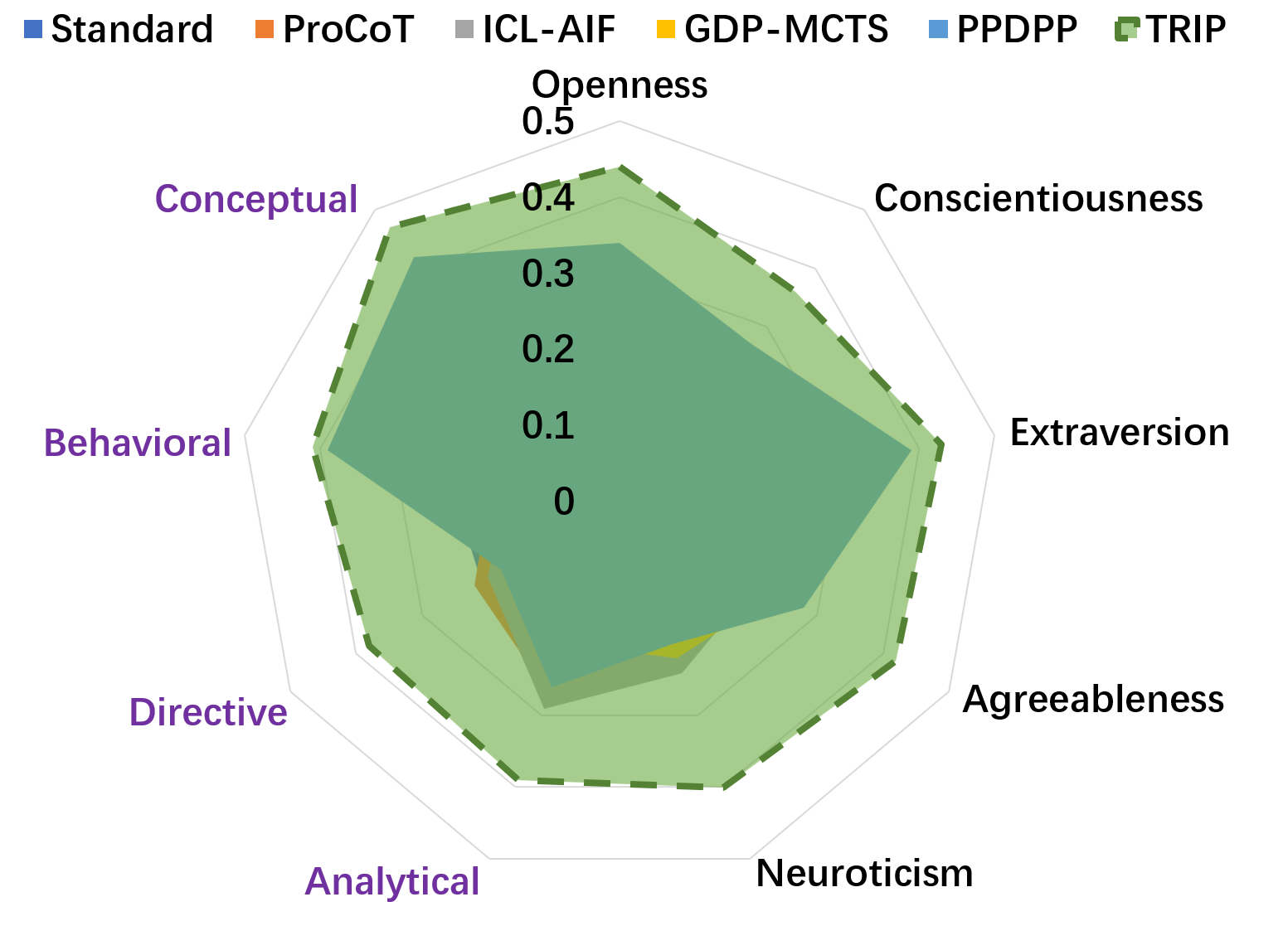}
\centering
\caption{The agents performance across various personas. We report their SL \% on the price negotiation task.
\textsc{Trip} achieves balanced improvements on all personas, significantly outperforming other agents by a considerable margin.}
\label{Radar_SL} 
\end{figure}

\begin{figure*}[ht] 
\includegraphics[width=\textwidth]{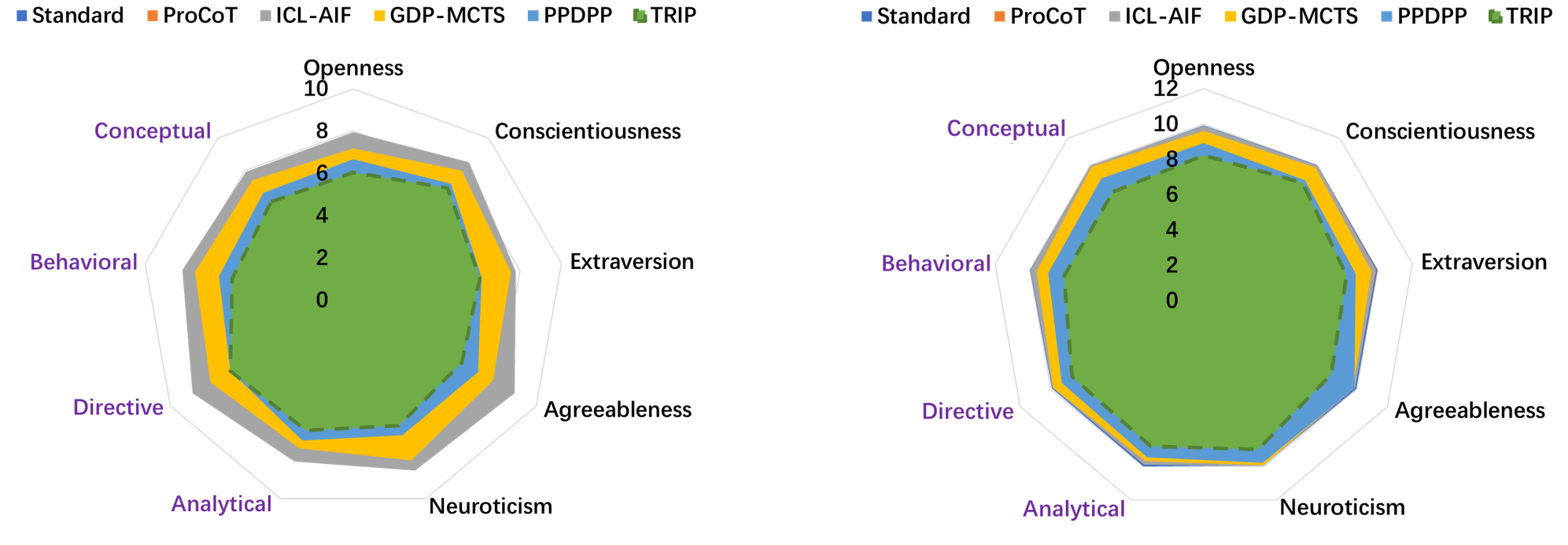}
\centering
\caption{The agents performance across various personas. We report their average turn on two tasks, namely price negotiation (\textit{Left}) and charity persuasion (\textit{Right}).
\textsc{Trip} achieves balanced improvements on all personas, significantly outperforming other agents by a considerable margin.}
\label{Radar_AT} 
\end{figure*}

In addition to the Success Rate, we report the agents performance across various personas using the metrics of Average Turn and Sale-to-List Ratio, as depicted in Figure \ref{Radar_AT} and Figure \ref{Radar_SL}.
We discover that the overall performance and analysis conclusions remain largely consistent with Section \ref{overaaa}.

\begin{table*}[h]
    \centering
    \begin{tabular}{l|p{9.5cm}}
        \toprule
        Dialogue Strategy & Explanation \\
        \hline
        Greetings & Please say hello or chat randomly. \\ \midrule
        Ask a question & Please ask any question about product, year, price, usage, etc. \\ \midrule
        Answer a question & Please provide information about the product, year, usage, etc. \\ \midrule
        Propose the first price & Please initiate a price or a price range for the product. \\ \midrule
        Propose a counter price & Please propose a new price or a new price range. \\ \midrule
        Use comparatives & Please propose a vague price by using comparatives with existing price.  \\ \midrule
        Confirm information & Please ask a question about the information to be confirmed. \\ \midrule
        Affirm confirmation & Please give an affirmative response to a confirm. \\ \midrule
        Deny confirmation & Please give a negative response to a confirm. \\ \midrule
        Agree with the proposal & Please agree with the proposed price. \\ \midrule
        Disagree with a proposal & Please disagree with the proposed price. \\ 
        \bottomrule
    \end{tabular}
    \caption{The negotiation strategies used in our \textsc{Trip} agent.}
    \label{tab:negotiate_strat}
\end{table*}

\begin{table*}[h]
    \centering
    \begin{tabular}{l|p{9.5cm}}
        \toprule
        Dialogue Strategy & Explanation \\
        \hline
         Logical Appeal  &  Please use of reasoning and evidence to convince the persuadee.    \\ \midrule
         Emotion Appeal  &  Please elicit the specific emotions to influence the persuadee.  \\ \midrule
         Credibility Appeal  &  Please use credentials and cite organizational impacts to establish credibility and earn the user’s trust. The information usually comes from an objective source (e.g., the organization’s website or other well-established websites).   \\ \midrule
         Foot in the Door  &  Please use the strategy of starting with small donation requests to facilitate compliance followed by larger requests.   \\ \midrule
         Self-Modeling  &  Please use the self-modeling strategy where you first indicates the persuadee own intention to donate and chooses to act as a role model for the persuadee to follow.   \\ \midrule
         Personal Story  &  Please use narrative exemplars to illustrate someone donation experiences or the beneficiaries positive outcomes, which can motivate others to follow the actions.   \\ \midrule
         Donation Information  &  Please provide specific information about the donation task, such as the donation procedure, donation range, etc. By providing detailed action guidance, this strategy can enhance the persuadee’s self-efficacy and facilitates behavior compliance.   \\ \midrule
         Source-related Inquiry  &  Please ask if the persuadee is aware of the organization (i.e., the source in our specific donation task).   \\ \midrule
         Task-related Inquiry  &  Please ask about the persuadee opinion and expectation related to the task, such as their interests in knowing more about the organization.   \\ \midrule
         Personal-related Inquiry  &  Please asks about the persuadee previous personal experiences relevant to charity donation.   \\ \bottomrule
    \end{tabular}
    \caption{The persuasion strategies used in our \textsc{Trip} agent.}
    \label{tab:persuasion_strat}
\end{table*}

\begin{table*}[!t]
    \centering
    \begin{tabular}{p{0.97\textwidth}}
    \toprule
    \textbf{\textit{The prompt for user persona generation}}     \\
    \midrule
    You need to select one attribute from each of the following persona types. \\
    ******** \\
    \textbf{Persona types} \\
    Big-Five Personality: ["openness", "conscientiousness", "extraversion", "agreeableness", "neuroticism"] \\
    Decision-Making Styles: ["directive", "analytical", "conceptual", "behavioral"] \\
    ******** \\
    Please generate a list of N fictional user profiles. \\
    \bottomrule
    \end{tabular}
    \caption{The prompt of user persona generation.}
    \label{prompt:persona generation}
\end{table*}

\begin{table*}[!t]
    \centering
    \begin{tabular}{p{0.97\textwidth}}
    \toprule
    \textbf{\textit{The prompt for user persona rephrase}}     \\
    \midrule
    You need to incorporate the following persona attributes and generate a cohesive persona description. \\
    You need to ensure the description is easy to understand. \\
    ******** \\
    Big-Five Personality: \\
    Decision-Making Style: \\
    ******** \\
    \midrule
    \textbf{An Example}: \\
    You are a 28-year-old female software developer. Your personality is characterized by openness to experience, which means you are curious, imaginative, and willing to try new things. In your occupation, you excel at analyzing problems and finding logical solutions. Your decision-making style is analytical, meaning you carefully consider all available information before making a choice.\\
    \bottomrule
    \end{tabular}
    \caption{The prompt of user persona rephrase.}
    \label{prompt:persona rephrase}
\end{table*}

\begin{table*}[!t]
    \centering
    \begin{tabular}{p{0.97\textwidth}}
    \toprule
    \textbf{\textit{The user simulator prompt for the price bargain task}}     \\
    \midrule
    Now enter the role-playing mode. In the following conversation, you will play as a seller in a price bargaining game. \\ \\
    Your persona: <Persona Description> \\ 
    You must follow the instructions below during chat.\\
    1. Your utterances and bargain behavior need to strictly follow your persona. Varying your wording and avoid repeating yourself verbatim! \\
    2. You can decide to change your target price flexibly based on your persona and the conversation.\\ \\
    
    Your Response Strategy: \\
    1. "Source Derogation": Attacks the other party or questions the item. \\
    2. "Counter Argument": Provides a non-personal argument/factual response to refute a previous claim or to justify a new claim. \\
    3. "Personal Choice": Provides a personal reason for disagreeing with the current situation or chooses to agree with the situation provided some specific condition is met. \\
    4. "Information Inquiry": Requests for clarification or asks additional information about the item or situation. \\
    5. "Self Pity": Provides a reason (meant to elicit sympathy) for disagreeing with the current terms. \\
    6. "Hesitance": Stalls for time and is hesitant to commit; specifically, they seek to further the conversation and provide a chance for the other party to make a better offer \\
    7. "Self-assertion": Asserts a new claim or refutes a previous claim with an air of finality/ confidence. \\
    8. "Others": Do not explicitly foil the negotiation attempts.
    \\ \\

    You are the seller who is trying to sell the \%s with the initial price of \%s. Product description: \%s. \\
    Please reply with only one short and succinct sentence. \\

    ******** \\ 
    Conversation History \\
    ******** \\
    \bottomrule
    \end{tabular}
    \caption{The comprehensive prompt of user simulators in the price negotiation task.}
    \label{prompt:comprehensive user simulator CB}
\end{table*}

\begin{table*}[!t]
    \centering
    \begin{tabular}{p{0.97\textwidth}}
    \toprule
    \textbf{\textit{The user simulator prompt for the charity persuasion task}}     \\
    \midrule
    Now enter the role-playing mode. In the following conversation, you will play as a Persuadee in a persuasion game. \\ \\
    Your persona: <Persona Description> \\ 
    You must follow the instructions below during chat.\\
    1. Your utterances need to strictly follow your persona. Varying your wording and avoid repeating yourself verbatim! \\
    2. Pretend you have little knowledge about the Save the Children charity. You have little willingness for donation at the beginning of conversation. \\
    3. Your willingness for donation depends on your persona and how influenced you are by the Persuader. \\
    4. You need to determine whether to donate money or not. If you think your are convinced by the Persuader, you should donate money for the charity. \\ \\
    
    Your Response Strategy: \\
    1. "Donate": show your willingness to donate. \\
    2. "Source Derogation": attacks or doubts the organisation's credibility. \\
    3. "Counter Argument": argues that the responsibility is not on them or refutes a previous statement. \\
    4. "Personal Choice": Attempts to saves face by asserting their personal preference such as their choice of charity and their choice of donation. \\
    5. "Information Inquiry": Ask for factual information about the organisation for clarification or as an attempt to stall. \\
    6. "Self Pity": Provides a self-centred reason for not being willing to donate at the moment. \\
    7. "Hesitance": Attempts to stall the conversation by either stating they would donate later or is currently unsure about donating. \\
    8. "Self-assertion": Explicitly refuses to donate without even providing a personal reason. \\
    9. "Others": Do not explicitly foil the persuasion attempts.\\ \\

    You are the Persuadee who is being persuaded by a Persuader. Please reply with only one short and succinct sentence. \\

    ******** \\ 
    Conversation History \\
    ******** \\
    \bottomrule
    \end{tabular}
    \caption{The comprehensive user simulator prompt for the charity persuasion task.}
    \label{prompt:comprehensive user simulator P4G}
\end{table*}

\begin{table*}[!t]
    \centering
    \begin{tabular}{p{0.97\textwidth}}
    \toprule
    \textbf{\textit{The Theory-of-Mind prompt for the price negotiation task}}     \\
    \midrule
    You are an expert in price bargain. \\
    Now give you a conversation history between a buyer and a seller, you need to infer the mental states and future actions of the seller.

    ******** \\ 
    Conversation History \\
    ******** \\
    \bottomrule
    \end{tabular}
    \caption{The ToM prompt for the price negotiation task.}
    \label{prompt:tom_CB}
\end{table*}

\begin{table*}[!t]
    \centering
    \begin{tabular}{p{0.97\textwidth}}
    \toprule
    \textbf{\textit{The Theory-of-Mind prompt for the charity persuasion task}}     \\
    \midrule
    You are an expert in charity persuasion. \\
    Now give you a conversation history between a persuader and a persuadee, you need to infer the mental states and future actions of the persuadee.

    ******** \\ 
    Conversation History \\
    ******** \\
    \bottomrule
    \end{tabular}
    \caption{The ToM prompt for the charity persuasion task.}
    \label{prompt:tom_P4G}
\end{table*}

\begin{table*}[!t]
    \centering
    \begin{tabular}{p{0.97\textwidth}}
    \toprule
    \textbf{\textit{The prompt of the ProCoT agent}}     \\
    \midrule
    \textbf{The Price Negotiation Task} \\
    Assume you are the buyer. Given the conversation history, in order to reach a better deal with the seller, please select the most appropriate dialogue strategy. \\
    You can only reply by selecting one of the following dialogue strategy to reach the goal: Greetings. Ask a question. Answer a question. Propose the first price. Propose a counter price. Use comparatives. Confirm information. Affirm confirmation. Deny confirmation. Agree with the proposal. Disagree with a proposal. \\
    The following is the conversation history: [conversation]\\
    \midrule
    \textbf{The Charity Persuasion Task} \\
    Assume you are the Persuader. Given the conversation history, in order to convince the persuadee to donate for charity, please select the most appropriate dialogue strategy. \\
    You can only reply by selecting one of the following dialogue strategy to reach the goal: Logical appeal, Emotion appeal, Credibility appeal, Foot in the door, Self-modeling, Personal story, Donation information, Source-related inquiry, Task-related inquiry, Personal-related inquiry. \\
    The following is the conversation history: [conversation]\\
    \bottomrule
    \end{tabular}
    \caption{The prompt design of the \textit{ProCoT} agent.}
    \label{prompt:procot_agent}
\end{table*}

\begin{table*}[!t]
    \centering
    \begin{tabular}{p{0.97\textwidth}}
    \toprule
    \textbf{\textit{The prompt of the ICL-AIF agent}}     \\
    \midrule
    \textbf{The Price Negotiation Task} \\
    Now enter the role-playing mode. In the following conversation, you will play as a coach in a bargain game. There will be a buyer and a seller bargaining about a product price. \\
    Your task is to read the conversation between the buyer and the seller, then provide suggestions to the buyer about how to buy the product with a lower price. Each suggestion should be only one short and succinct sentence. \\
    The following is the conversation: [conversation] \\
    Question: What are your suggestions? Answer:\\
    \midrule
    \textbf{The Charity Persuasion Task} \\
    Now enter the role-playing mode. In the following conversation, you will play as a coach in a persuasion game. There will be a persuader who is trying to persuade a persuadee for charity donation. \\
    Your task is to read the conversation between the persuader and the persuadee, then provide suggestions to the persuader about how to convince the persuadee to make a donation. Each suggestion should be only one short and succinct sentence. \\
    The following is the conversation: [conversation] \\
    Question: What are your suggestions? Answer:\\
    \bottomrule
    \end{tabular}
    \caption{The prompt design of the \textit{ICL-AIF} agent.}
    \label{prompt:icl_aif_agent}
\end{table*}

\begin{table*}[!t]
    \centering
    \begin{tabular}{p{0.97\textwidth}}
    \toprule
    \textbf{\textit{The prompt of our TRIP agent in price negotiation}}\\
    \midrule
    Now enter the role-playing mode. In the following conversation, you will play as a buyer in a price bargaining game. \\
    You are the buyer who is trying to buy the \%s with the price of \%s. Product description: \%s \\
    Please reply with only one short and succinct sentence. [action] Now start the game.\\
    \bottomrule
    \end{tabular}
    \caption{The prompt design of the \textsc{Trip} agent for price negotiation.}
    \label{prompt:trip_agent_CB}
\end{table*}

\begin{table*}[!t]
    \centering
    \begin{tabular}{p{0.97\textwidth}}
    \toprule
    \textbf{\textit{The prompt of our TRIP agent in charity persuasion}}\\
    \midrule
    Now enter the role-playing mode. In the following conversation, you will play as a Persuader who is trying to persuade the Persuadee to donate to the charity called Save the Children. \\
    Save the Children is head-quartered in London, and they work to help fight poverty around the world. Children need help in developing countries and war zones. Small donations like \textdollar1 or \textdollar2 go a long way to help. \\
    You are the Persuader who is trying to convince the Persuadee to donate to a charity called Save the Children. [action] \\
    Please reply with only one short and persuasive sentence.\\
    \bottomrule
    \end{tabular}
    \caption{The prompt design of the \textsc{Trip} agent for charity persuasion.}
    \label{prompt:trip_agent_P4G}
\end{table*}

\end{document}